\documentclass[graybox]{svmult}

\usepackage{mathptmx}       
\usepackage{helvet}         
\usepackage{courier}        
\usepackage{type1cm}        
\usepackage{makeidx}         
\usepackage{graphicx}        
\usepackage{multicol}        
\usepackage[bottom]{footmisc}

\usepackage{amsmath}
\usepackage{cite}
\usepackage{url}
\usepackage{tikz}             
\usepackage{tikz-qtree}
\usetikzlibrary{matrix}
\usepackage{float}
\usepackage{algorithm}
\usepackage{algpseudocode}

\usepackage{subfigure}
\usepackage{caption}
\usepackage{subcaption}

\usepackage{todonotes}

\usepackage[defaultlines=3,all]{nowidow}

\usepackage{xcolor}
\usepackage{listings}
\makeindex

\begin{document}



  

\title{Crafting Generative Art through Genetic Improvement: Managing Creative Outputs in Diverse Fitness Landscapes}
\titlerunning{Crafting Generative Art through Genetic Improvement}

\author{Erik M. Fredericks, Denton Bobeldyk, and Jared M. Moore}

\institute{Erik M. Fredericks \at College of Computing, Grand Valley State University, USA, \email{frederer@gvsu.edu}
\and
Denton Bobeldyk \at College of Computing, Grand Valley State University, USA, \email{bobeldyd@gvsu.edu}
\and
Jared M. Moore \at College of Computing, Grand Valley State University, USA, \email{moorejar@gvsu.edu}
}

\maketitle

\abstract{
%
Generative art is a rules-driven approach to creating artistic outputs in various mediums.
For example, a fluid simulation can govern the flow of colored pixels across a digital display or a rectangle placement algorithm can yield a Mondrian-style painting.
Previously, we investigated how genetic improvement, a sub-field of genetic programming, can  automatically create and optimize generative art drawing programs. 
One challenge of applying genetic improvement to generative art is defining fitness functions and their interaction in a many-objective evolutionary algorithm such as Lexicase selection.  
%
%
Here, we assess the impact of each fitness function in terms of the their individual effects on generated images, characteristics of generated programs, and impact of bloat on this specific domain. 
Furthermore, we have added an additional fitness function that uses a classifier for mimicking a human's assessment as to whether an output is considered as ``art.'' 
This classifier is trained on a dataset of input images resembling the glitch art aesthetic that we aim to create.
Our experimental results show that with few fitness functions, individual generative techniques sweep across populations.  
Moreover, we found that compositions tended to be driven by one technique with our current fitness functions. 
%
%
Lastly, we show that our classifier is best suited for filtering out noisy images, ideally leading towards more outputs relevant to user preference. 
}


\section{Introduction}
\label{sec:intro}
Generative art produces an output
using a rules-based approach, algorithm, or other heuristic~\cite{boden.2009,forbes.2013,cabral.1993,liu.2007,smedt.2011,fredericks.ause.2024,fredericks.gi.2023}.  
Example outputs include digital images/animations, models intended for 3-D printing, and vector images intended for pen plotting.
It can be used to teach programming concepts to students~\cite{greenberg.2007,bergstrom.2015,peppler.2005,shiffman.2012,dehouche.2023}, visualize trends in datasets~\cite{comba.2020.data}, or illustrate the behaviors of complex algorithms~\cite{boden.2009,forbes.2013,cabral.1993,liu.2007,smedt.2011}.
We follow the definition of generative art proposed by Boden and Edmunds~\cite{boden.2009}, however we acknowledge that alternative definitions exist.
Generative art is highly-amenable to genetic programming (GP) as it can comprise one or more drawing functions that result in substantially different outputs, depending on the parameterization.
For this research, we use grammatical evolution (GE), a subset of GP, for evolving programs using an input grammar constraining the solution space.

Computer-based generative art may rely upon one or more parameterized drawing functions to create an output, where function parameters are used to induce variety in the output each time the function is invoked~\cite{boden.2009,forbes.2013,cabral.1993,greenberg.2007,shiffman.2012}.  
For example, the behavior and output of a flow field (c.f., Section~\ref{sec:bg}) may be varied by changing how the underlying vector field is defined (e.g., octaves and falloff in the case of a Perlin noise-based field~\cite{perlin.2001}), whether the drawing mechanism will be connected lines or individual particles, and the color palette used when drawing.  
Additionally, the digital representation of the flow field could be realized as a physical artistic piece by using a pen plotter or 3D printer.
Each technique and associated parameters can be encoded within a genome and result in a drawing program that can be evolved with GP.
However, assessing the ``goodness'' of a solution is a non-trivial problem as preference is difficult to mathematically quantify and is individually-subjective.
We use GE for constraining the structure of evolved programs with a grammar~\cite{smedt.2011,ryan.1998}. 
This approach can be considered as a form of genetic improvement (GI), or automatically improving programs using evolution~\cite{langdon.2020}.

Assessing the individual and combined impact of the fitness functions in an open-ended, many-objective evolutionary process is an ongoing and challenging problem.
Our prior research efforts (\texttt{GenerativeGI}) aimed to create outputs that resemble glitchy or corrupted images for aesthetic purposes~\cite{fredericks.ause.2024,fredericks.gi.2023}.
However, automatically assessing the quality of those outputs is non-trivial and determining if a generated image is ``art'' is subjective, necessitating a human examiner.  
Further complicating assessment are the different interpretations of computational aesthetics as noted by Greenfield~\cite{greenfield2005origins}.
Our primary goal with \texttt{GenerativeGI} is to automate the assessment process suitable for GP. 
As human preference is difficult to mathematically quantify, we use proxy functions (e.g., maximizing the novelty of the generated program, incorporating as many techniques as possible, maximizing pixel difference between outputs, etc.) to attempt to emulate the preferences of the designer. 
%
%
As the number of fitness functions grows to incorporate various dimensions of artistic evaluation, the individual influence of each fitness function is not immediately evident.  
This chapter investigates the impact of the fitness objectives on the overall evolutionary process when creating generative art.

We perform a combinatorial analysis of fitness objectives within the evolutionary selection process to determine the impact and contribution of each to the overall outputs.
The generated programs and outputs are evaluated in terms of their relevance to the desired aesthetic style of outputs, fitness values, and execution metrics. 
Additionally, we include a new fitness function that uses a machine learning classifier, trained on an abstract art corpus~\cite{kaggle.genart.dataset.2024}, to both provide additional evolutionary pressure and serve as a post-processing mechanism for an ``art vs. not art'' evaluation. 
Our experimental results indicate that fewer fitness functions result in generative art techniques sweeping across populations, leading to very similar outputs.
However, we also note that with all of our fitness functions included that we still see sweeping occurring, indicating that other fitness functions,  evolutionary pressures, or diversity preservation options are required for further novelty.
Additionally, we found that \texttt{GenerativeGI} tended to evolve solutions that were skewed towards a dominant generative technique per experimental configuration.
%
We also found, as is somewhat expected, that program bloat was occurring in our outputs as well.  
Bloat is potentially useful in this domain, although it also can be a challenge. 
Additional drawing functions can lead to interesting/unexpected outputs at the expense of the time required to render the techniques.
%
Lastly, we found that our classifier tends to classify ``noisy'' outputs as ``not art,'' ideally leading to a better evolutionary feedback mechanism for moving away from glitch art.
For archival purposes we include our full datasets on Zenodo\footnote{See \url{https://zenodo.org/records/12796744}.} and updated codebase on Github.\footnote{See \url{https://github.com/efredericks/GenerativeGI/tree/GPTP-args}.}
%

The remainder of this chapter is organized as follows.  Section~\ref{sec:bg} overviews generative art, GE and GI, our classifier, and related work.
Section~\ref{sec:expr} then presents and discusses our investigation into fitness function impact on \texttt{GenerativeGI}.
Finally, Section~\ref{sec:discussion} summarizes our results and presents future directions for research.


\section{Generative Art, Genetic Improvement, and Convolutional Neural Network Classifiers}
\label{sec:bg}
This section provides relevant information on generative art, GE and GI, and our convolutional neural network (CNN) classifier.  We additionally present related work in this section.

\subsection{Generative Art}
\label{sec:art}
Generative art is typically considered to be a process for creating an artistic output (either digitally or physically) via a heuristic or algorithm~\cite{boden.2009}.\footnote{This differs significantly from the recent explosion of interest in outputs created via generative adversarial networks, though the terms are similar.}
Famous generative artists include Joshua Davis, Zack Lieberman, Vera Molnár, Frieder Nake, Casey Reas, Manfred Mohr, Lillian Schwartz, Georg Nees, Roman Verostko, Rafaël Rozendaal, and Sol LeWitt.
Moreover, generative art can be considered as an umbrella term for multiple forms of creative outputs, including artistic expression, visualizations of datasets or algorithms, and creative coding purposes for teaching programming concepts~\cite{boden.2009,forbes.2013,cabral.1993,liu.2007,smedt.2011,fredericks.ause.2024,fredericks.gi.2023,comba.2020.data}.
Any form of visualization can serve as the basis of a generative art program. 
We briefly introduce two examples to illustrate the process: flow fields and circle packing.\footnote{A larger number of examples included within \texttt{GenerativeGI} are described in detail in \cite{fredericks.ause.2024}.}

\begin{figure}
\centering  
\subfigure[Flow field example.]{\includegraphics[width=0.30\linewidth]{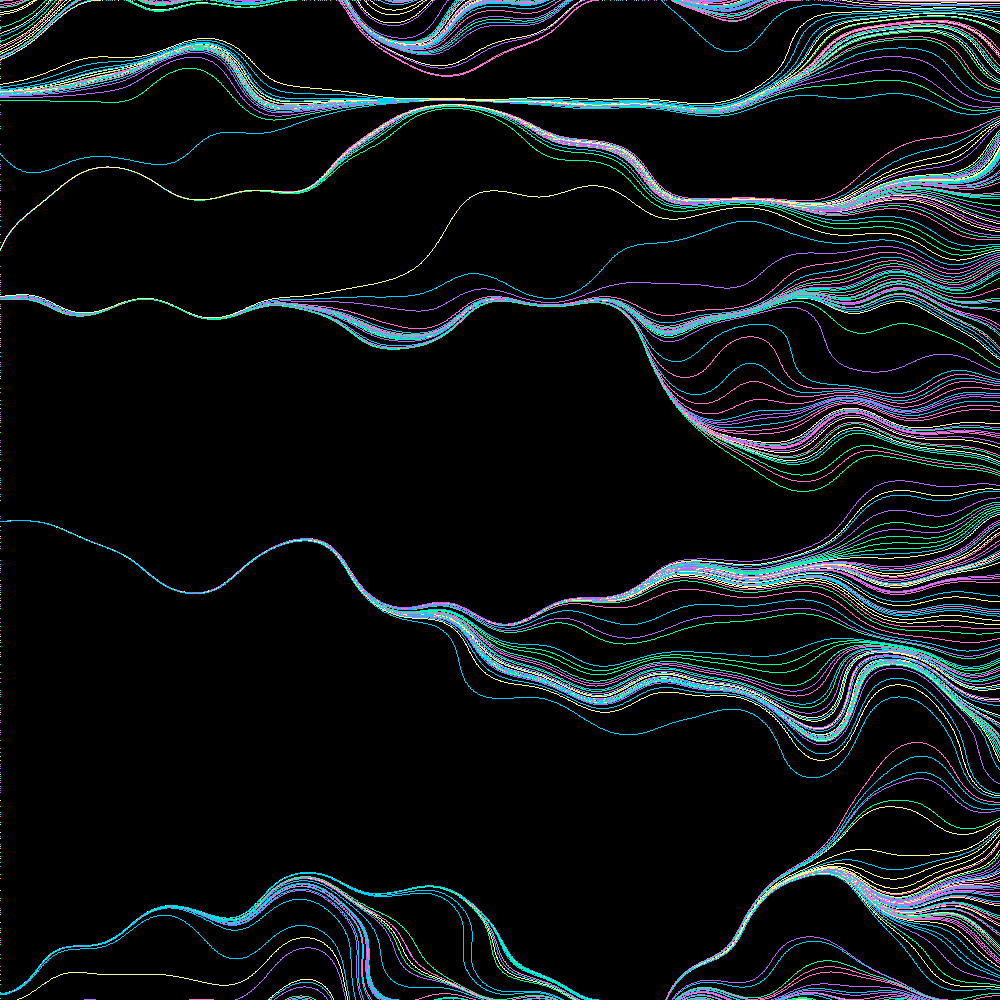}}
\subfigure[Circle packing example.]{\includegraphics[width=0.30\linewidth]{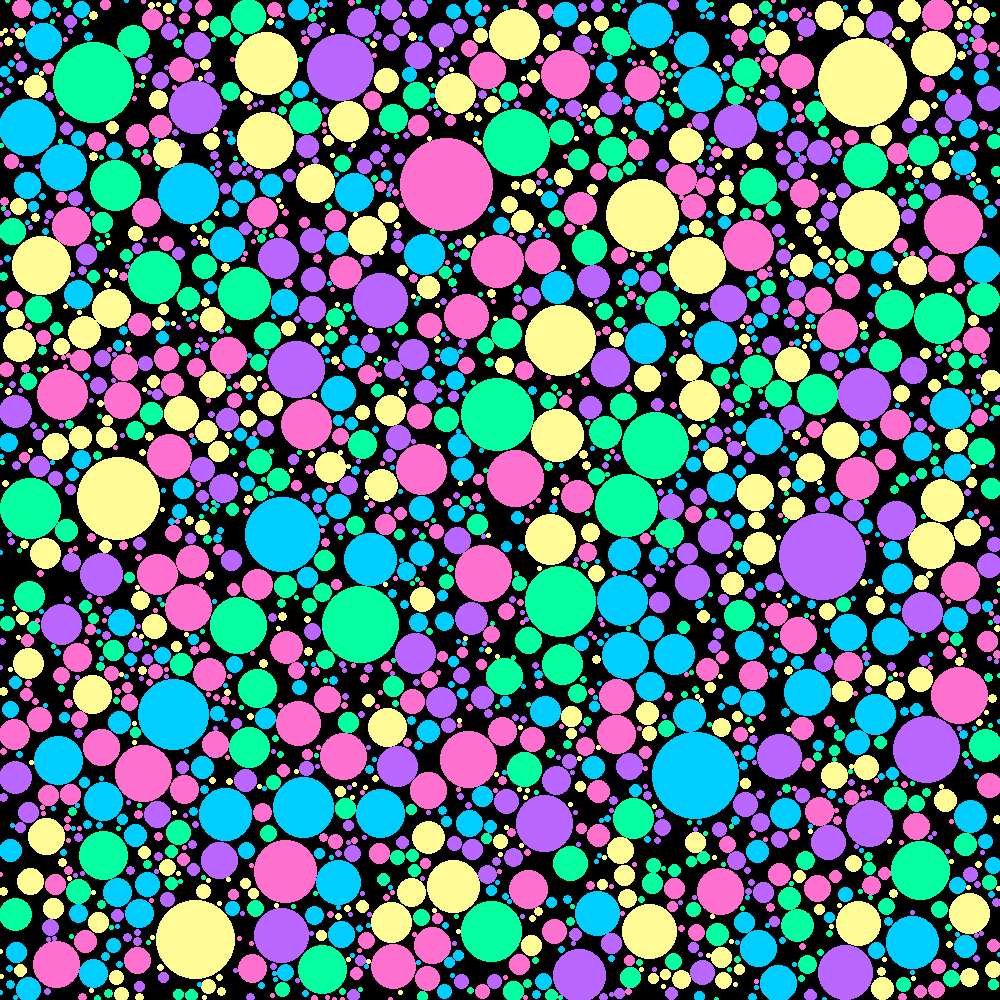}}
\subfigure[Combined flow field / circle packing example.]{\includegraphics[width=0.30\linewidth]{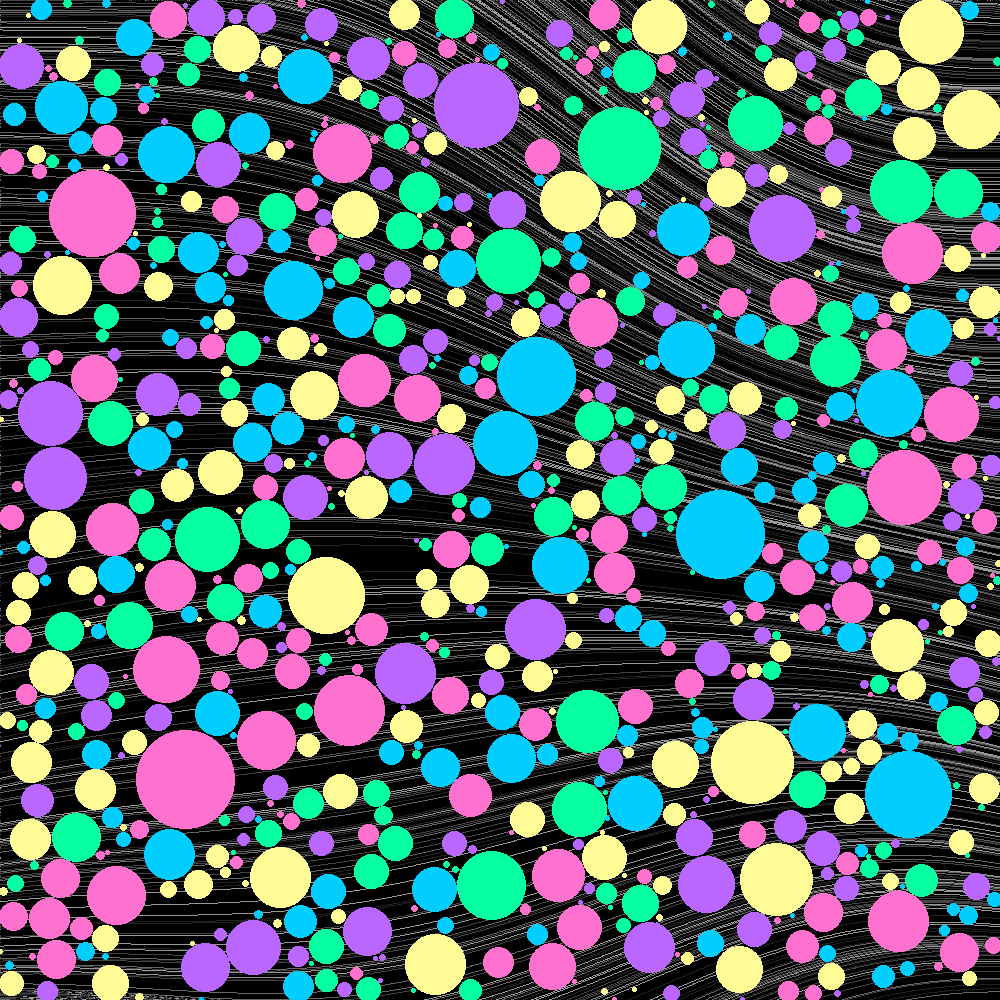}}

\vspace{-0.15in}

\caption{Sample generative art outputs.}
\label{fig:genart-samples}
\end{figure}

\textbf{Flow fields} visualize motion along vector fields, typically in terms of velocity or orientation~\cite{cabral.1993}.
Figure~\ref{fig:genart-samples}(a) presents a sample flow field output using particle-driven motion across a Perlin noise-mapped vector field~\cite{perlin.2001}.
In this output a noise value (using Perlin noise) is calculated based on $x$ and $y$ coordinates within the image canvas and that value is then mapped to an angle on [$-2.0 * \pi$, $2.0 * \pi$].
A particle (comprising a position and color) will draw a point on the canvas with its specified color at its current position.
The particle will then sample the underlying noise field and calculate its next position based on the corresponding angle.

\textbf{Circle packing} aims to fit as many circles as possible into a given space while requiring that no circle overlaps with another.
Figure~\ref{fig:genart-samples}(b) presents a sample circle packing output.



While each algorithm alone can create interesting artistic outputs, the complexity of the output can significantly increase when techniques are combined.
Figure~\ref{fig:genart-samples}(c) demonstrates the output when a flow field is run and then followed by circle packing (with a different color palette for differentiation).



A combination of artistic techniques constitutes a drawing program where the invocation and ordering of art functions, along with their respective parameter instantiation(s), varies from execution to execution.
Such a process is amenable to evolutionary computation for optimization in selection of the drawing techniques, the configuration of parameters for each, and the order in which they are executed~\cite{sims1993interactive,baluja1994towards}.

Note that evolutionary computation has been used to support numerous creative endeavors, including \cite{sims1993interactive,baluja1994towards,correia2013evolving,Correia2019,machado2008experiments,machado2012improving,clune2011evolving,cluneendlessforms,denHeijer.2013,smedt.2011,johnson.2016}.
An index of papers published at the Conference on Computational Intelligence in Music, Sound, Art and Design (EvoMUSART) from 2003 onwards provides a comprehensive list of highly-related papers to this domain~\cite{evomusart}.


In our prior work, \texttt{GenerativeGI}, we used GE as the means for evolving drawing programs within a constrained solution space~\cite{fredericks.ause.2024,fredericks.gi.2023}.
Genomes within \texttt{GenerativeGI} are defined as a combination of parameterized drawing techniques expressed as a string, where we used single-point crossover and single-point mutation.
Mutation will either select a new drawing technique with new parameters at the mutation point, update parameters for the existing technique at the mutation point, or shuffle the order of all individual techniques within the genome when activated.

Large-language model-based techniques may leverage GPT-3~\cite{floridi.2020} and GPT-4~\cite{openai2023gpt4,phung2023generative} to support synthesis using massive datasets of existing samples.
Such an approach is typically governed by user prompts for creating relevant outputs, with techniques including diffusion models~\cite{zhang2023text} and generative adversarial networks (GANS)~\cite{goodfellow2020generative}.
Conversely, our approach relies on developer-provided drawing techniques and evolutionary search to discover ideal drawing programs.
Of note is that there are significant concerns regarding image copyright and ownership for GAN/diffusion model training datasets~\cite{balan2023ekila}.


\subsection{Grammatical Evolution and Genetic Improvement}
\label{sec:gi}

GE is a form of GP in that the genome is represented as a variable-length byte list and parsed by a grammar (typically in Backus-Naur form), as opposed to a tree or stack-based representation~\cite{koza.1994, smedt.2011, ryan.1998}.
This genome places additional constraints on generated outputs as the only options that are feasible are syntactically-correct to the grammar, ideally reducing bloat and/or invalid programs.
However, such an approach requires a significant amount of domain expertise to properly construct the grammar.

GI extends the concepts of GP/GE by improving the source code and/or parameters of a generated program~\cite{langdon.2014,harman.2015.gi4gi,petke.2014,petke.2017,langdon.2017.genetic}.
Examples of GI include automated bug repair via software patches~\cite{legoues.2012.systematic}, repurposing old functions for new program uses~\cite{langdon.2017.genetic}, and reducing program energy consumption via optimized code processes~\cite{bruce.2015}.

\texttt{GenerativeGI}~\cite{fredericks.ause.2024,fredericks.gi.2023} uses GE and GI to automatically create parameterized sequences of drawing functions that yield artistic outputs, where our outputs are governed by a many-objective selection mechanism (i.e., Lexicase selection~\cite{Spector2012}) to enable search in a large solution space.
Here, selection of individuals for crossover/mutation occurs by comparing individual fitness objectives in a randomized order.
If an individual performs better in the compared objective then it will be immediately selected for reproduction.
If a tie occurs then the selection process moves on to the next objective.
\texttt{GenerativeGI} uses both integer and real-valued objectives.
Comparison in real-valued objectives is enabled by $\epsilon$-Lexicase selection for ranking individuals within a specified range~\cite{LaCava2016}, where our specified range is set to $\epsilon$ $=$ $0.85$ based upon empirical evidence.

Related to Lexicase selection are multi-objective optimization approaches, such as NSGA-II~\cite{deb.2002}.
While pareto-based approaches are powerful for multiple objectives, we intend to continue adding new fitness functions to the point where pareto-based approaches may begin to break down.

Regardless, fitness functions are typically defined with some objective in mind.
There seems to be little existing work on teasing apart the impact of individual fitness functions on the evolutionary process as a whole, instead relying on domain expertise to determine if a fitness function is ``good.'' 
Previously, Panait and Luke investigated how competitive fitness (i.e., one- or multi-population individuals working cooperatively or competitively towards a common goal, similar to co-evolution) can impact evolutionary outcomes~\cite{panait.2002}. 
Such an approach can be an interesting future direction for \texttt{GenerativeGI}, as currently it makes use of traditional, single-population individuals.
Deb and Gupta previously investigated knee-points within Pareto-fronts in multi-objective optimization as points of potential trade-off optimization~\cite{deb2011understanding}.
In terms of our research, knee points may indicate similar tradeoffs for our fitness functions, however they do not necessarily indicate the usefulness of the fitness function itself.

\subsection{Art Classification}
\label{sec:classifier}
Human preference is challenging to mathematically quantify, yet there was a critical need to include a technique for measuring the extent to which an image resembles ``art'' that would facilitate gradual improvements in successive iterations towards the ``art'' category. 
Integrating a human reviewer into the evolutionary loop would be infeasible 
as each of the reviews would be time consuming and add an exponential amount of time to each run. 
As such, machine learning could provide a proxy for human preference when trained on a relevant dataset.

A CNN is a type of machine learning model that is capable of learning and then predicting to what class an image belongs~\cite{lecun2015deep}. 
The model is trained on images that belong to each class and learns how to extract relevant features that are used for class prediction. 
Output is typically a numeric value associated with how confident the model feels an image belongs to a certain class,  
where the class that has the largest numeric value is selected by the model. 
To integrate this model properly into our pipeline the feature extractor component was separated from the classifier, 
where generated images were then analyzed to detect if they contain features found by the model.

\begin{figure}[htb!]
\centering  
\subfigure[Typical CNN including classification layers.]{\includegraphics[width=0.52\linewidth]{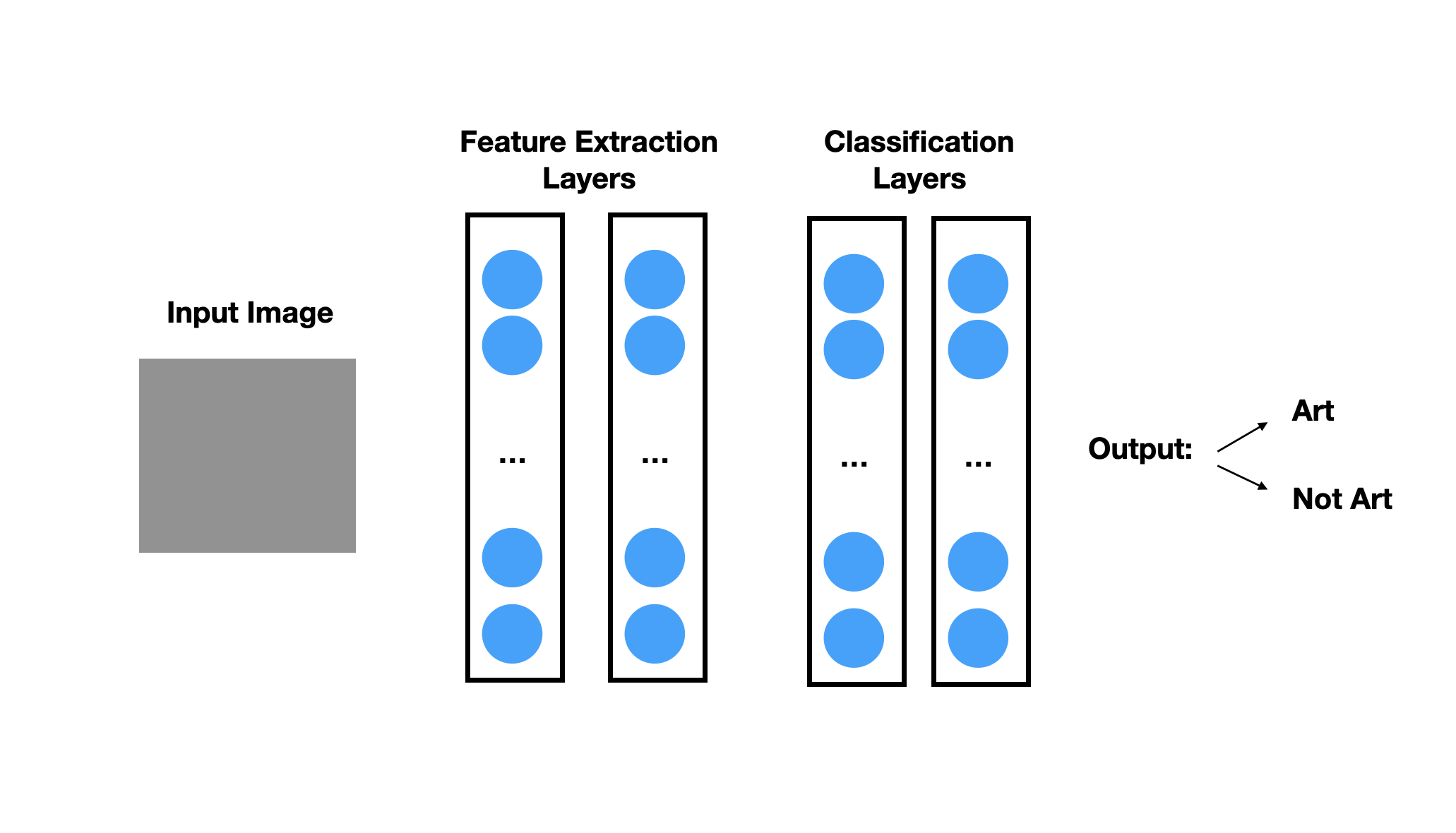}}
\subfigure[Revised CNN for use in evolutionary algorithm.]{\includegraphics[width=0.42\linewidth]{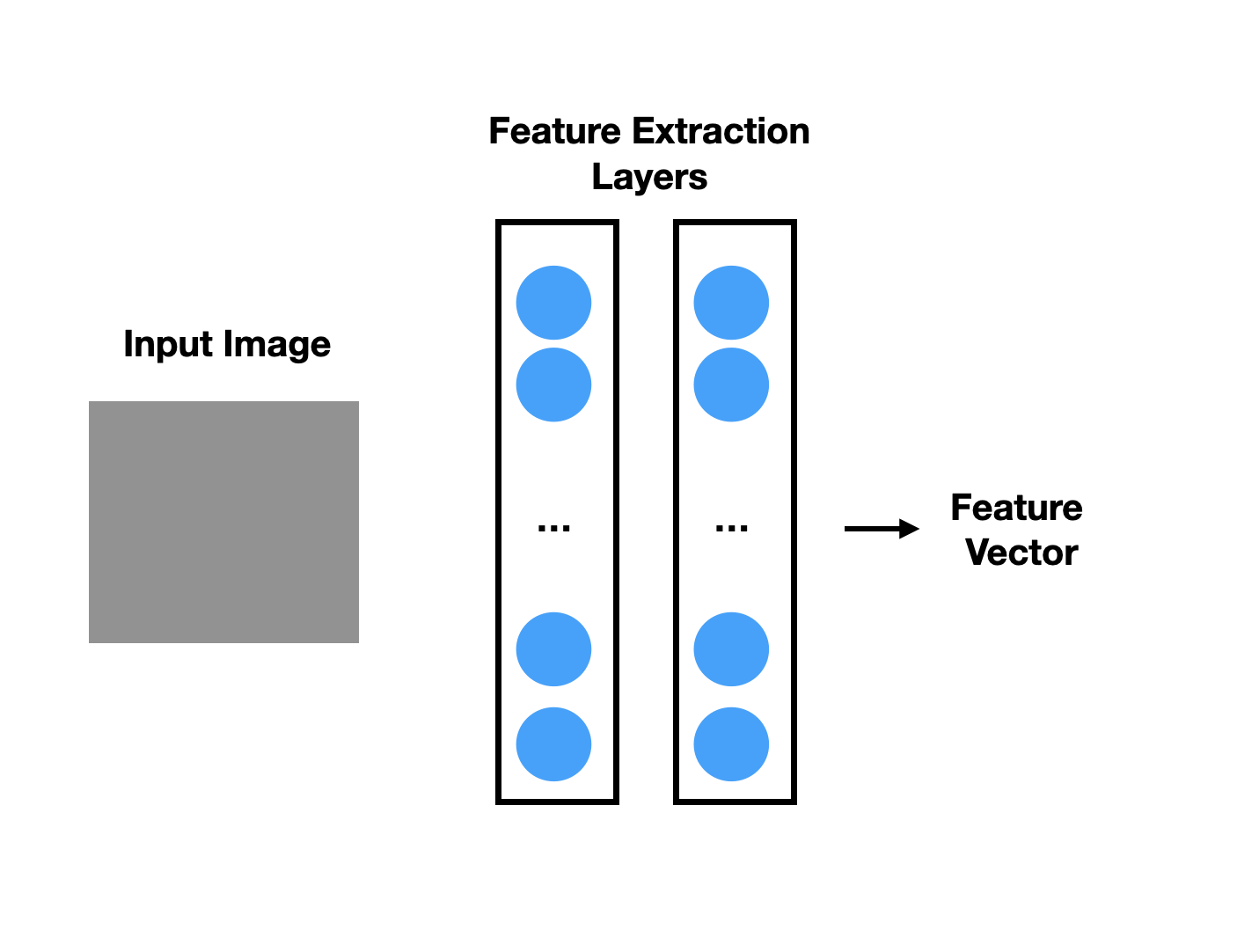}}
\caption{CNN visualizations.}
\label{fig:cnns}
\end{figure}

A standard CNN model (see Figure~\ref{fig:cnns}(a)) typically contains a configurable amount of layers, where the first layers are responsible for extracting features and the latter layers are responsible for classifying learned features. 
The classification layers generally take a high dimensional vector and reduce those layers to a smaller dimensional vector (i.e., the output layer). 
Our CNN implementation removes the typical classification layers to generate a vector of learned features (see Figure~\ref{fig:cnns}(b)) for use within the evolutionary loop.
The generated feature vector can measure how close an image is to either the ``art'' or ``not art'' feature space. 

We used a dataset of abstract art~\cite{kaggle.genart.dataset.2024} and randomly-generated noise images to determine the ``art'' and ``not art'' feature spaces, respectively.
A standard CNN model was trained on the dataset in an attempt to learn the features of the ``art'' class. 
The classification layers were then removed after training and each of the art images were applied as input to the model. 
The mean of the output feature vectors generated from all of the ``art'' class images was calculated as well as those from ``not art.'' 
The revised model accepts a new input image and outputs a fitness score that could be used to determine how close the image was to the ``art'' class.


A potential challenge is that classifiers have been proven to be easily fooled into mislabeling outputs in both artistic endeavors and other classification tasks~\cite{correia2013evolving,Correia2019,machado2012improving,nguyen2015deep}.
Such results indicate that classifiers are not always generalizable and such must be recognized when including a classifier as part of the feedback process.
For our purposes we use our CNN as a companion objective within the selection process and do not fully rely on it for quantification.
As is described in Section~\ref{sec:expr}, our classifier is helpful in reducing noisy or blurry outputs and can help the designer focus on specific images within a large number of outputs.





\section{Lexicase Selection, Fitness Landscapes, and Examining Search Criteria}
\label{sec:expr}




Generally, fitness functions are defined with developer domain knowledge and an aim to optimize a specific aspect of a possible solution.  
Depending on the domain, crafting fitness functions can be straightforward, translating standard problem metrics into a fitness objective.  
For example, a fitness function for a robot learning to walk might maximize absolute distance traveled from its origin.
When moving to multiple objectives, however, Pareto fronts and competing objectives~\cite{deb.2002,deb2011understanding} can introduce complexity, obscuring individual fitness function contributions.
The impact and usefulness of fitness functions can be non-trivial to assess when the problem domain is open-ended or domain knowledge is lacking. 
For example, an optimal solution in a generative art application might be guided by any number of fitness functions as each output can induce a different ``feature'' crafting a desirable composition.
Such features could include favoring brush strokes over pixel placement, maximizing negative space in the canvas to provide ``breathing room'' for drawings, or maintaining a consistent color palette for aesthetic purposes. 

In our prior efforts~\cite{fredericks.ause.2024,fredericks.gi.2023} we included fitness functions that were intended to induce a ``glitch art'' effect (i.e., in the style of aesthetically-broken or corrupted images).
We introduced a number of fitness functions (c.f., Table~\ref{table:ffs}) that guided search towards glitch art while attempting to maximize diversity in the population of images.
However, these techniques often resulted in images that appeared noisy and cluttered.
Our application domain is fundamentally a problem of human preference that must be mathematically quantified, as incorporating a human in the loop would be impractical given that full evaluations can span multiple days of processing time within a high-performance computing (HPC) environment.
To this end, our fitness functions served as proxies for human preference. \texttt{GenerativeGI} included the fitness functions as described within Table~\ref{table:ffs}, except $FF_{ac}$, which is a new component introduced for this chapter.


\begin{table}[htb!]
\centering
\begin{tabular}{||c | l ||} 
 \hline
$FF_{pc}$ & (Maximize) Pairwise comparison (RMS pixel difference between images) \\ \hline
$FF_{gc}$ & (Minimize) Number of duplicate genes\\ \hline
$FF_{ut}$ & (Maximize) Diversity of unique techniques\\ \hline
$FF_{cd}$ & (Maximize) Chebyshev difference between images\\ \hline
$FF_{ns}$ & (Minimize) Negative space (target 70\%)\\ \hline
$FF_{ac}$ & (Minimize) Art classifier (numerical ``art'' or ``not art'')\\ 
& \textit{Lower values indicate closer relation to ``art''-classified images} \\ \hline
\end{tabular}
\caption{Fitness functions used by \texttt{GenerativeGI}. $FF_{pc}$ through $FF_{ns}$ were included in \cite{fredericks.ause.2024}.}
\label{table:ffs}
\end{table}

\noindent \textbf{Experiments}.  In this work, we performed a combinatorial analysis of our fitness objectives through a ``leave $x$ out'' approach to determine the relative impact of each objective.
Depending on the experimental configuration, objectives were toggled on or off during the selection process, however each were still measured during evaluation for post-run processing.
We denote the fitness objectives actively used in selection in each experimental configuration ($EC$) with subscripts.
For example, a configuration with negative space ($ns$) and the art classifier ($ac$) active is denoted as: $EC_{ns, ac}$.

$15$ replicates per experimental configuration were run.  
Each replicate comprised $100$ individuals evolving for $100$ generations.
Crossover and mutation rate were set to $0.5$ and $0.4$, respectively.
%
An individual is created via single-point crossover with genetic material from two parents with a $50\%$ chance, otherwise it is cloned.
Mutation is then performed on an individual with a $40\%$ chance using single-point mutation.



{
\textbf{Time analysis}. 
We first evaluated individual technique timing prior to our investigation into fitness as each of our submitted HPC jobs would fail with a wall time error (a maximum of 88 hours was specified for each individual replicate).
We then set a time limit of three minutes per evaluation for each individual, after which evaluation would cease regardless of genome length.
This change enabled all experimental configurations and replicates to complete within a significantly shorter amount of time.

The issue with timeouts can be seen as an effect of program bloat as our genome length was quite large and each drawing technique requires time to execute.  
Therefore, we analyzed the times of each of our generative art techniques individually.  
As can be seen from Figure~\ref{fig:technique-times}, one particular technique (\texttt{flow-field}) takes significantly longer than other techniques to run.  
In reviewing grammar outputs from cancelled runs we noted that the \texttt{flow-field} was highly prevalent in the population with some individuals having multiple flow fields in their genome.  
As we have a second version of the flow field algorithm (i.e., \texttt{flow-field-2}), we opted to remove \texttt{flow-field} and re-run our experiments with an increased timeout of four minutes per evaluation.

\begin{figure}[htb!]
\centering
\includegraphics[width=4.6in]{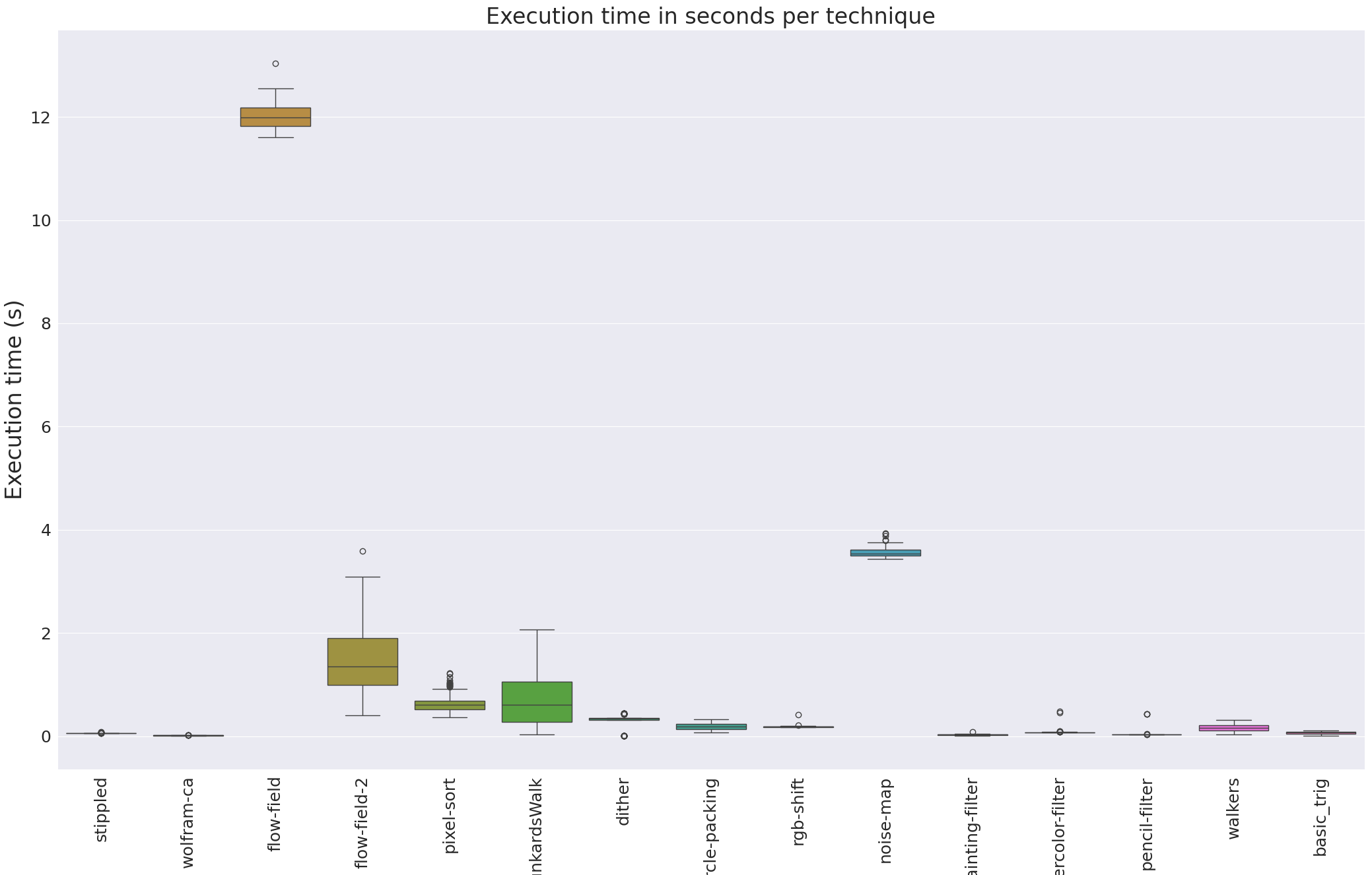}

\vspace{-0.13in}

\caption{Total time required for each individual technique across 100 randomly-instantiated invocations.}

\label{fig:technique-times}
\end{figure}

Generative art techniques can be computationally expensive requiring calculation at the individual pixel level or evaluating a high number of individual agents used to create an artifact.  
Combining multiple techniques together can yield significant increases in runtime that make evolutionary processes prohibitively expensive from a wall clock perspective.  
Still, maintaining a variety of techniques in an individual's genome enhances the possibility of a novel combination of genes to yield ``interesting'' results.  
To balance these competing challenges, we implemented a maximum evaluation time (three minutes) for an individual to express its genome in an image and re-ran our experimental replicates for analysis.  
If a composition was still being generated at three minutes, the currently active technique was allowed to complete and the remaining genes were not expressed.  

Next, we visually inspected the outputs from the final generation of each experimental configuration by generating an image collage.
One of the most striking observations was the amount of techniques sweeping across populations, depending on the number of fitness functions included.
For example, Figure~\ref{fig:collage}(a) shows a collage of outputs from $EC_{ut,ac}$ (i.e., $FF_{ut}$ and $FF_{ac}$). 
Even though a fitness objective was activated to maximize the number of unique techniques, visually it appears that the population converges on a common technique (i.e., \texttt{basic-trig}, or drawing trigonometric functions) as well as a common color palette. 
Whereas, outputs from the initial population from this configuration were quite diverse. 



\begin{figure}
\centering  
\subfigure[Collage of final outputs from one replicate of $EC_{ut,ac}$.]{\includegraphics[width=0.48\linewidth]{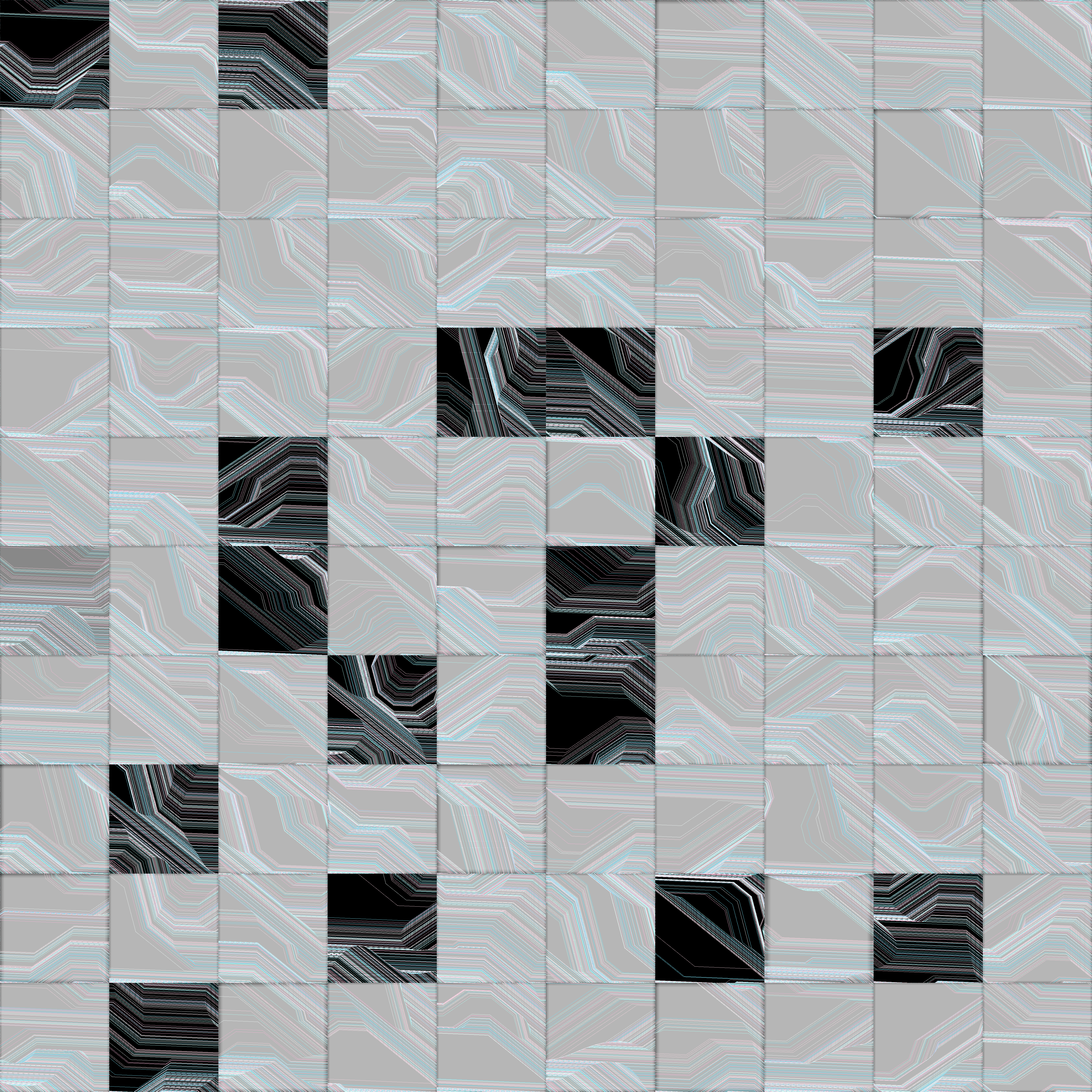}}
\subfigure[Collage of final outputs from one replicate of $EC_{pc,gc,ut,cd,ns,ac}$.]{\includegraphics[width=0.48\linewidth]{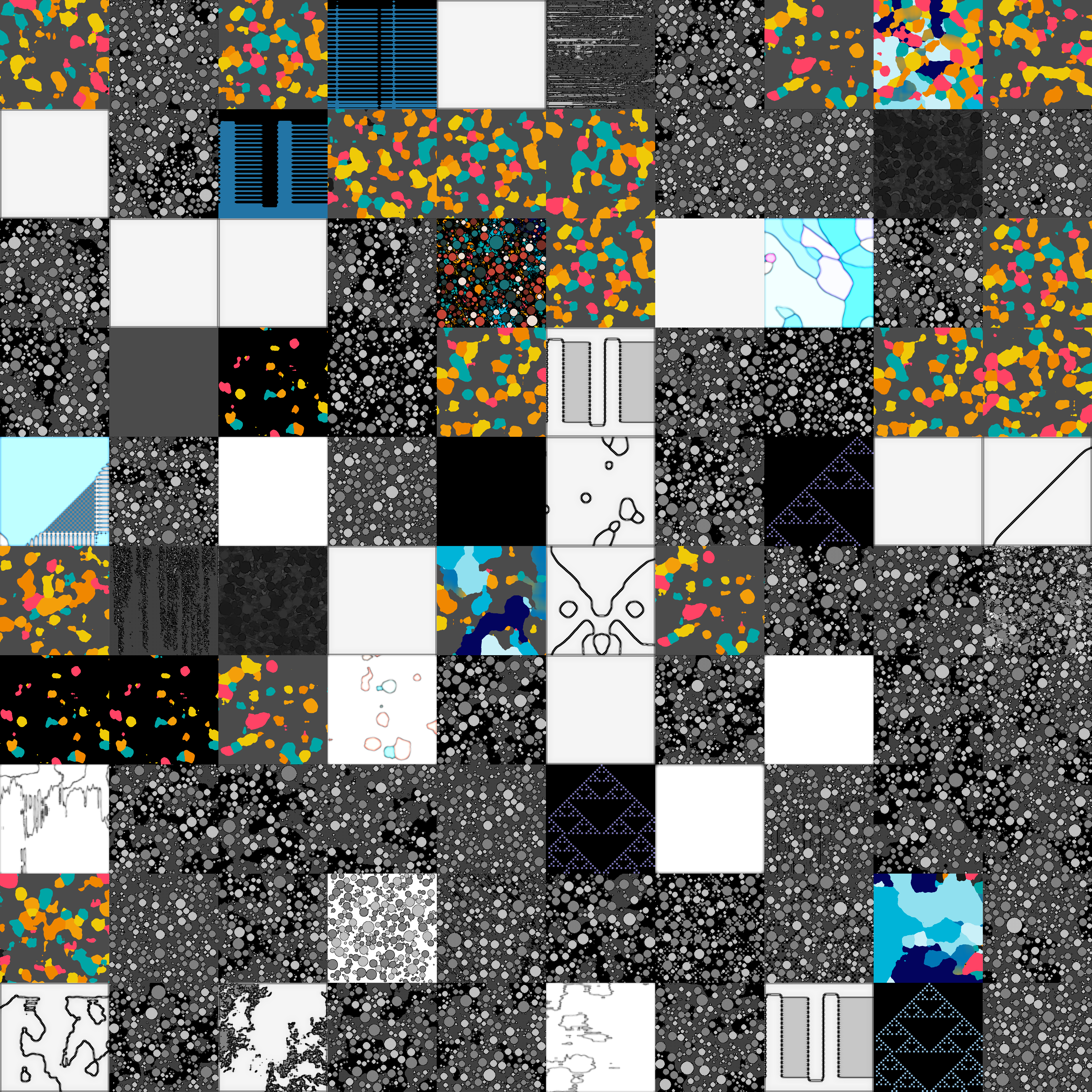}}

\vspace{-0.15in}

\caption{Sample collages of final outputs from $EC_{ut,ac}$ and $EC_{pc,gc,ut,cd}$ from one replicate.  Each collage comprises the outputs from the final evolved population to demonstrate the difference between the different objectives used for selection.}
\label{fig:collage}
\end{figure}

Conversely, Figure~\ref{fig:collage}(b) demonstrates a run with more fitness objectives active during selection.
Techniques are still clearly sweeping across the population, however there exists more variety in the outputs.
This difference is observed across replicates suggesting that activating more fitness objectives may lead to niching in the population.


We then examined the fitness values across each configuration.
Although only a subset of fitness functions were active in selection, all were calculated per individual for post-run analysis.
Figure~\ref{fig:heatmap2} shows a heatmap comprising each experimental configuration and fitness objectives. 
Each fitness function returns values on varying ranges, and as such, we have normalized each to a range of [$0.0$, $1.0$], where $1.0$ is considered most fit and $0.0$ is considered least fit.

All experimental configurations minimize the number of duplicate genes shown by the dark red for the $FF_{gc}$ row.  
However, many of the other fitness functions are maximized only when they are used in active selection for specific experimental configurations.  
This behavior is shown in the geometric patterns that emerge horizontally for each $FF$ row.  
Of note, $FF_{ac}$ remains relatively high across runs suggesting that most experimental configurations produce ``art'' as determined by the CNN classifier (and colloquially by the authors).  
Negative space, another common concept in composition, only reaches high scores~(near $70\%$ in an image) when actively selected for.  
While the classifier may be indicating ``art'' this suggests that negative space is not being strongly considered by the $FF_{ac}$ classifier as $FF_{ns}$ occasionally has a low score.  


\begin{figure}[htb!]
\centering
\includegraphics[width=5.0in]{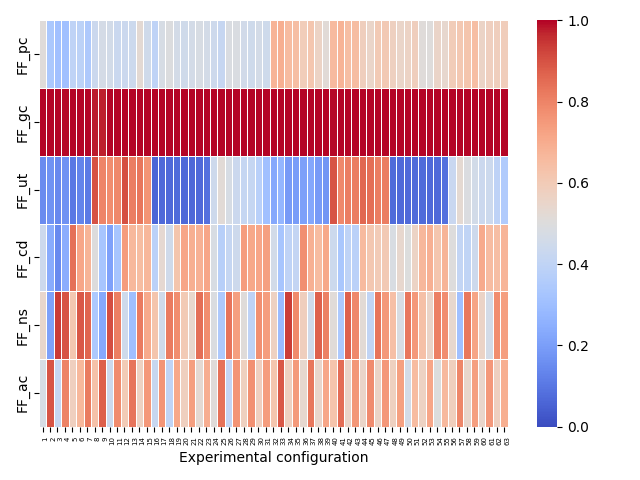}

\vspace{-0.15in}

\caption{Heatmap of normalized and averaged fitness values for final population per experimental configuration (three minute evaluation timeout, \texttt{flow-field} removed). Experimental configuration numbers correlate to a truth table of activated fitness objectives.}
\label{fig:heatmap2}
\end{figure}

\noindent \textbf{Output Classification}. 
In our prior work, the evolved compositions often resembled ``glitch art,'' our target style. 
However, they exhibited artifacts such as blurriness or noise not seen in human-crafted glitch art.  
One possible reason for this is that each fitness function is assessing specific aspects of a composition such as duplicate genes, a target negative space, or maximizing difference between images in a population.  
Considering the whole of an image in a composition is not amenable for these fitness functions.  
Here, we added the previously mentioned CNN classifier as $FF_{ac}$ to try and assess images more holistically. 
After conclusion of the runs, we ran our classifier on the final population of images for all experimental configurations to determine the total number of generated images classified as most resembling our training dataset. 
This analysis serves two purposes: to ensure that $FF_{ac}$ is functioning as expected within our evolutionary process and to provide the authors with additional information for deeper visual inspection on particular images.
For example, an image classified as ``not art'' may look similar to others in a collage view, however zoomed in they tend to look noisy, cluttered, or blurry.
Figure~\ref{fig:art-comparison}(a) shows an example of an output image classified as ``art'' and Figure~\ref{fig:art-comparison}(b) shows an example of a ``not art'' image.  
As can be seen, the images are visually similar and yet the ``not art'' image is significantly blurrier, leading to the supposition that the classifier is providing feedback on the validity of generated images. 
For all experimental configurations, the number of images classified as ``art'' significantly outnumber the number classified as ``not art.''
These results suggest that further refinements/additions to our classifier are necessary to aid the search process.


\begin{figure}[htb!]
\centering  
\subfigure[``Art''-classified image.]{\includegraphics[width=0.48\linewidth]{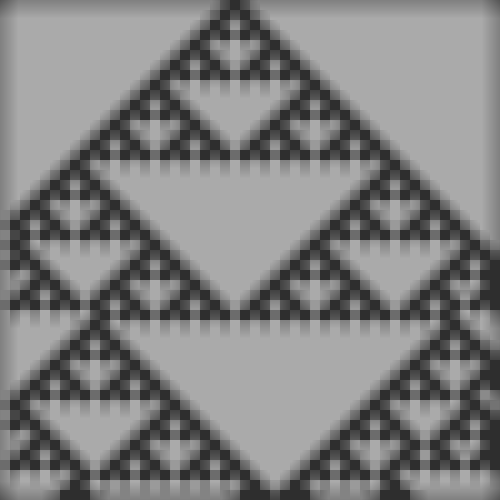}}
\subfigure[``Not art''-classified image.]{\includegraphics[width=0.48\linewidth]{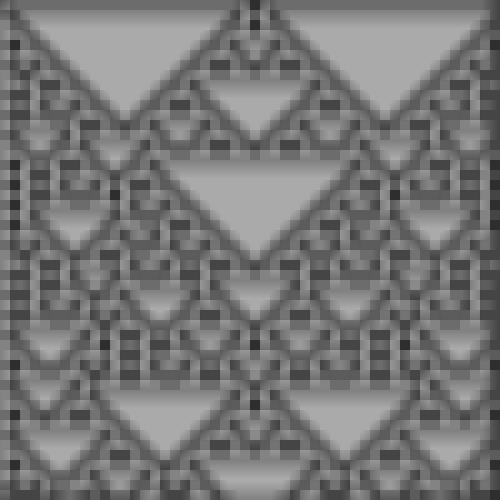}}

\vspace{-0.15in}

\caption{Classified ``art'' vs. ``not art'' images from $EC_{pc,gc,ut,cd}$.}
\label{fig:art-comparison}
\end{figure}

\section{Discussion} 
\label{sec:discussion}
This chapter has presented an empirical analysis of the impact of fitness objectives within the context of an open-ended optimization problem where human preference supersedes quantitative metrics.
Specifically, we investigated how different combinations of fitness functions can influence generative art outputs, given that our defined fitness functions were used in an ad hoc fashion~\cite{fredericks.ause.2024,fredericks.gi.2023}.
For this particular domain, results suggest that the number of included fitness objectives during selection can have an impact on artistic techniques sweeping across the population.
Moreover, some fitness objectives (i.e., $FF_{gc}$, $FF_{ac}$) tended to score high throughout all configurations whereas some fitness objectives (i.e., $FF_ns$) only scored high when active.
Finally, we found that our classifier is well-suited for filtering images that do not meet basic qualifications for being considered ``art'' (e.g., noisy or blurry images).

Future directions for this path of research include investigating different selection mechanisms (e.g., Dalex~\cite{ni2024dalex}, MAP-ELITES~\cite{mouret.2015.illuminating}) to focus on diversity of results, improving the underlying data structure managing our generative art techniques to enable interactions between individual techniques, and investigating different machine learning classifiers for assessing output quality.

\begin{acknowledgement}
The authors gratefully acknowledge the reviewers' comments which helped to improve the manuscript.
We gratefully acknowledge support from the Michigan Space Grant Consortium (award number 80NSSC20M0124), resources from the Grand Valley State University High Performance Computing cluster (which is supported by the Academic Research Computing department), and Grand Valley State University.  The authors additionally thank Alexander Lalejini and Abigail Diller for their inputs as well.
\end{acknowledgement}

\bibliographystyle{spmpsci}
\bibliography{efredericks_master}

\end{document}